%% file: emnlp2023.tex
\pdfoutput=1
\documentclass[11pt]{article}

\usepackage{EMNLP2023}

\usepackage{times}
\usepackage{latexsym}
\usepackage{graphicx}

\usepackage[T1]{fontenc}
\usepackage[utf8]{inputenc}

\usepackage{microtype}

\usepackage{inconsolata}

\usepackage{xspace}

\newcommand\ie[0]{i.e.,\xspace}

\usepackage{amsmath}
\usepackage{amssymb}
\usepackage{booktabs}

\title{Asking More Informative Questions for Grounded Retrieval}

\author{Sedrick Keh \\
  Carnegie Mellon University\\
  \texttt{skeh@alumni.cmu.edu} \\\And
  Justin T. Chiu \\
  Cornell Tech \\
  \texttt{jtc257@cornell.edu} \\\And
  Daniel Fried \\
  Carnegie Mellon University \\
  \texttt{dfried@cs.cmu.edu}
  }

\begin{document}

\maketitle

\input{sections/0.abstract}
\input{sections/1.intro}
\input{sections/2.background}

\input{sections/3.methods}
\input{sections/4.experimental-setup}
\input{sections/5.results-analysis}

\input{sections/6.discussion}
\input{sections/7.related-work}
\input{sections/8.conclusion}

\input{sections/acknowledgements.tex}

\input{sections/limitations}
\input{sections/ethics}
\bibliography{anthology,custom}
\bibliographystyle{acl_natbib}

\appendix
\input{sections/appendix.tex}

\end{document}

%% file: sections/0.abstract.tex
\begin{abstract}
When a model is trying to gather information in an interactive setting, it benefits from asking informative questions. However, in the case of a grounded multi-turn image identification task, previous studies have been constrained to polar yes/no questions \cite{white-etal-2021-open}, limiting how much information the model can gain in a single turn. We present an approach that formulates more informative, open-ended questions. In doing so, we discover that off-the-shelf visual question answering (VQA) models often make presupposition errors, which standard information gain question selection methods fail to account for. To address this issue, we propose a method that can incorporate presupposition handling into both question selection and belief updates. Specifically, we use a two-stage process, where the model first filters out images which are irrelevant to a given question, then updates its beliefs about which image the user intends. Through self-play and human evaluations, we show that our method is successful in asking informative open-ended questions, increasing accuracy over the past state-of-the-art by 14\%, while resulting in 48\% more efficient games in human evaluations. 
\end{abstract}

%% file: sections/1.intro.tex
\section{Introduction}
\label{sec:introduction}
As NLP models are increasingly deployed in interactive settings, it is key that these models are able to gather information about the user's intentions and the underlying world context. Models might do this by asking questions to the user --- however, asking informative questions is challenging, as it relies on reasoning about the current context, the history of the interaction, and potential future plans. Past work on question generation for interaction has formulated questions that are predicted to be informative, but has typically used questions with constrained answer spaces \cite{rao-daume-iii-2018-learning,yu-etal-2020-interactive,white-etal-2021-open}, which can lead to inaccurate and inefficient interactions in contextually rich tasks. 

\begin{figure}
    \centering
    \includegraphics[width=\columnwidth]{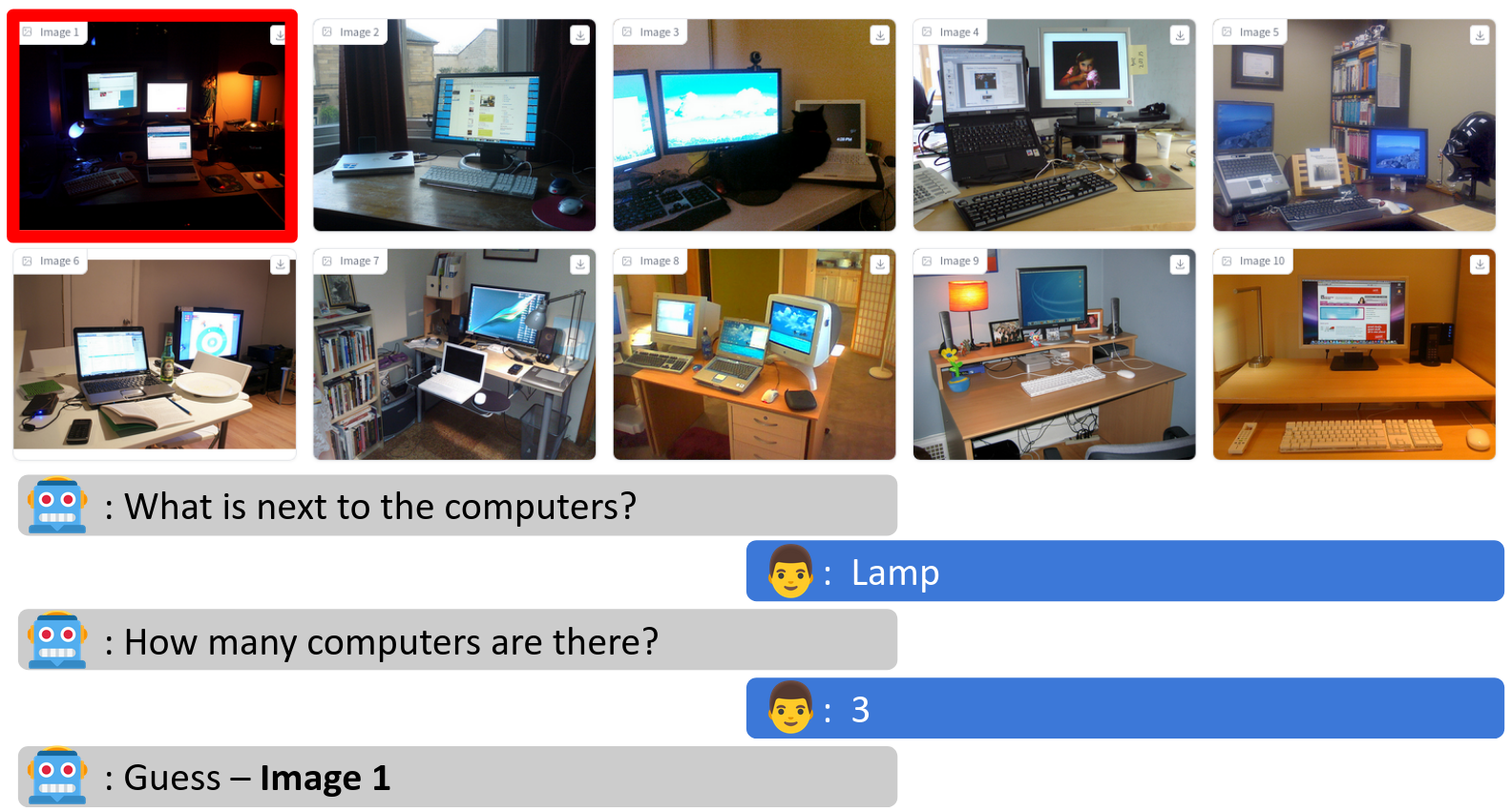}
    \caption{We propose a method for interactive image identification, where our model's goal is to ask the most informative questions to quickly and accurately guess the target image (highlighted in red).
    \vspace{-2em}
    }
    \label{fig:game}
\end{figure}

We present an approach for generating open-ended questions in one such contextually rich task: a multi-turn image identification task~\cite{white-etal-2021-open} inspired by interactive retrieval. In this task, a model is presented with an array of images, one of which is a \emph{target} image known only to a human user. The model has to identify this image by formulating questions for the user to answer. A sample game can be seen in \autoref{fig:game}, and more details can be found in Section \ref{sec:background}. We select this task as a challenging test bed for strategic contextual interaction: requiring strong abilities in grounding (to distinguish between similar images that differ only in their less-salient details, e.g. all images in \autoref{fig:game} contain a computer) and planning (asking informative questions that take into account the history of interaction).

Previous approaches to this task \cite{white-etal-2021-open} have used polar yes/no questions in the form of "IsA" questions. 
Although these IsA questions are formulated to be maximally informative according to an expected information gain metric (Section \ref{subsec:expected-information-gain}), polar questions do not scale well to settings where images are similar, or with a larger number of images. It is also not straightforward to simply replace IsA questions with broader open-ended questions (i.e. \emph{wh-} questions such as who, where, when, etc.) 
We show (Section \ref{sec:results-and-analysis}) that doing so naively produces questions with presupposition errors, where an off-the-shelf visual question answering (VQA) model gives confident answers to questions that are irrelevant to an image. 
An example of a presupposition error would be asking "What is the dog eating?" when there is no dog in the target image. 
This failure renders the standard information gain metric inappropriate and results in a chain of errors in belief updates over the model's target images. 
Constraining to polar yes/no questions largely protected previous approaches from facing such issues, but once open-ended questions are introduced, the chances for presupposition errors rise substantially.

We propose a method that can generate open-ended informative questions in the interactive image retrieval setting, relaxing the polar yes/no constraints, while at the same time being able to handle presuppositions. 
We do this by conducting a two-stage process: first explicitly checking for presuppositions to filter out irrelevant images, then calculating information gain based only on the relevant images. We show that after handling such errors, asking open-ended \emph{wh-} questions substantially outperforms constraining to polar yes/no questions. In our human evaluations, our method results in 48\% more efficient games compared to the past state-of-the-art for this task~\cite{white-etal-2021-open}, while improving accuracy by 13\%. We further conduct ablation studies to verify that our two-stage pipeline is indeed successful in eliminating presupposition errors, and that it can generalize to settings where there are substantially more images to choose between. Our method is even more successful in this more challenging setting, resulting in an average of 3.5 times shorter games.

%% file: sections/2.background.tex
\section{Background}
\label{sec:background}
We discuss the task and notation used, as well as some key concepts/tools used in our model.

\subsection{Task}
\label{subsec:task}
The multi-turn grounded image retrieval task was formally introduced in \citet{white-etal-2021-open}, though it can be viewed as an extended setting of a Lewis signaling game \cite{lewis1969}. In this setting, there are a total of $k$ images (we consider $k=10$ and $k=100$) containing one target image, chosen at random. There are two agents who can both observe all the images. One agent (the responder) knows the identity of the target image, while the other agent (the guesser) does not. The goal of the guesser is to ask clarifying questions in order to accurately identify the target image in as few turns as possible. This dialogue proceeds for multiple turns, until either the guesser is confident enough to make a guess, or until a set threshold (number of turns) is reached, in which case the guesser will have to make a guess. In our paper, the guesser is a model (whose goal is to identify the target image by asking clarifying questions), while the responder can either be a human or another model. 

More formally, we define $I$ = $\{i_1, i_2, \dots i_k\}$ as the set of images, and $y$ as the target image. In each turn $t$ of the interaction, the guesser (model) asks a question $q$, and the responder answers with a response $r$. We additionally define $x^t$ as the history of the interaction $(q_1, r_1, q_2, r_2, \dots q_t, r_t)$.

In our version of the task, we add a new component --- the "No Answer" option. Instead of responding to the guesser's clarifying question, the responder may opt to not provide an answer. The option to deliberately not answer a question will provide valuable information to a model.
Using our notation, we represent this as $r_{null}$. This simple change is a necessary addition if we wish to incorporate \emph{wh-} questions, as the responder needs an appropriate way to respond to the guesser's question if it does not apply to the target image. 

\subsection{Expected Information Gain}
\label{subsec:expected-information-gain}
To quantify the most informative question, one common metric \citep{lindley,rao-daume-iii-2018-learning,yu-etal-2020-interactive,white-etal-2021-open} is the expected information gain of a given question $\mathrm{EIG}(y,r;q,x^{t})$. A model maintains a \emph{belief distribution} $P(y\mid x^t, q, r)$ over which image is the target, and aims to ask questions that will minimize uncertainty in this belief distribution, taking answer likelihood into account. This is given by minimizing the conditional entropy of the belief distribution, in expectation over possible answers to the question: 
\begin{equation}\arg\min_{q} 
\label{eq1}
\mathop{\mathbb{E}}_{p(y \mid x^t)} \mathop{\mathbb{E}}_{p(r \mid q,y)} [-\ln P(y \mid x^t,q,r)]
\end{equation} 
Note that this requires a model $p(r\mid q, y)$ to predict the user's response to any $(q, I, y)$ set. For this, we use a proxy VQA answering model, which we discuss more in Section \ref{subsec:top-q-selection}.

%% file: sections/3.methods.tex
\begin{figure*}[t]
    \centering
    \includegraphics[width=\textwidth]{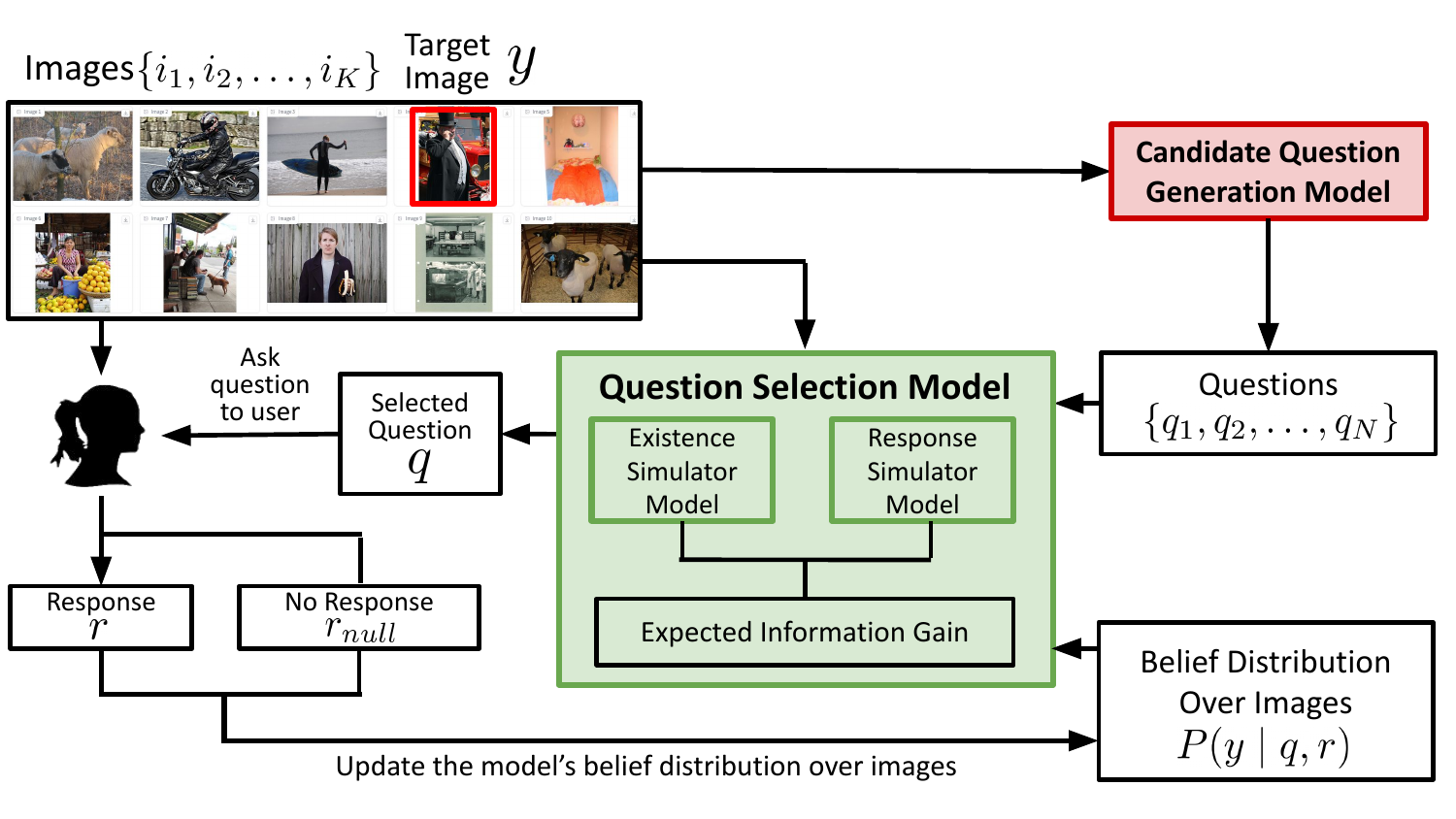}
    \caption{Overview of our pipeline/flow for a single turn of the interactive image retrieval task. The colored boxes represent modules (which can be a model, a ground truth oracle, or a human player), while the clear boxes represent various inputs/outputs during the game.
    \vspace{-1em}
    }
    \label{fig:pipeline}
\end{figure*}

\section{Methods}
\label{sec:methods}
Our main pipeline is illustrated in Figure \ref{fig:pipeline}. We use a training-free approach, using only off-the-shelf pre-trained VQA and text processing models. Using the image contexts, we first generate a large pool of possible candidate questions $Q$ (Section \ref{subsec:q-gen}), then select the question $q \in Q$ which gives the highest expected information gain about the belief distribution (which estimates which image is the target), taking into consideration the various presupposition assumptions (Section \ref{subsec:top-q-selection}). Upon asking the question $q$ and receiving a response $r$, the model then updates its belief distribution accordingly (Section \ref{subsec:belief-updates}).

\subsection{Candidate Question Generation}
\label{subsec:q-gen}
In order to generate appropriate questions, we first generate a list of captions, then convert each caption to a list of several questions, as was done in \citet{white-etal-2021-open}. This approach allows us to leverage the strong capabilities of image captions combined with text generation models (for caption-to-question), which results in more diverse and higher quality questions as compared to directly conditioning on the image.

More formally, for each image $i \in I$, in order to generate question set $Q_i$, we first obtain a caption $C_i$, then convert the caption $C_i$ into $\{q_{i_1}, q_{i_2}, \dots q_{i_j}\} \in Q_i$,  where each $q_{i_j}$ represents a possible way to convert caption $C_i$ into a question. For instance, if the caption $C_i$ is ``A man brushing his teeth with a toothbrush'', then possible questions $q_{i_1}, q_{i_2}, q_{i_3}$ could be ``What is the man doing?'', ``Where is the man?'', and ``What color is the toothbrush?'' Notably, all our questions are open-ended \emph{wh-} questions, as opposed to the polar yes/no questions used in previous studies~\citep{white-etal-2021-open}. To convert each caption $C_i$ into a set of candidate questions $Q_i$, we leverage in-context learning capabilities of large language models such as GPT-4 \cite{gpt4}. 
Specifically, we provide a system prompt describing the task in detail, as well as 4 few-shot examples in the format "Caption:[$C_i$]. Questions:[$q_{i_1}, q_{i_2}, \dots q_{i_j}$]". In order to make the in-context learning more robust, we vary $j$ across the different captions, ranging from 3 to 7, with an average of 4.75 questions per caption. More details on the prompting process and our exact prompts can be found in \autoref{appendix:prompts}. 

\subsection{Top Question Selection}
\label{subsec:top-q-selection}
Selecting the most informative question is done based on highest expected information gain (Eq. \ref{eq1}). In order to calculate expected information gain, we use a response simulator model (i.e., a VQA model) which allows us to calculate the probability $P(r \mid q, i)$, for any $(r,q,i)$ tuple. Intuitively, this probability represents how likely a user would be to respond to a question in a given way, if a particular image is the target.

However, as mentioned earlier, the model often faces presupposition errors, in which case the probability updates may not accurately reflect the true amount of information gained. As such, rather than simply using the expected information gain, we modify the question selection process to account for presuppositions. To do this, we employ a two-step process, as denoted by the Existence Simulator Model and the Response Simulator Model in \autoref{fig:pipeline}. 
We verify the importance of each of these components in ablations in Section \ref{subsec:ablation-presupposition-handling}.

\textbf{Existence Simulation Step:} The goal of this step is to identify which images a particular questions is relevant to, as it would not make sense to run response simulation on images where the question does not apply.

For each question, we define $\text{Relevance}(q,i)$ as the indicator variable that is $1$ when question $q$ is relevant to image $i$, and $0$ otherwise. We compute $P(\text{Relevance}(q,i))$ as follows: (1) We convert wh- question $q$ into a set of polar yes/no questions $\{q_{yn_1}, q_{yn_2}, \dots q_{yn_j}\}$. This is done by parsing the question with a constituency 
parser \cite{kitaev-klein-2018-constituency} and extracting all possible subjects (nouns or noun phrases). Each possible subject $s_1, s_2, \dots s_j$ is then directly converted into a yes/no question $q_{yn_1}, q_{yn_2}, \dots q_{yn_j}$ by asking "Is there a $s_j$?" or "Are there $s_j$?" (2) Using our VQA model, we implicitly ask each of these yes/no questions and take the mean across questions, so $P(\text{Relevance}(q,i))$ is computed as 
\begin{equation}
\label{eq2}
\frac{1}{j} \sum_{k=1}^j VQA(r_\text{yes} \mid q_{yn_k}, i)
\end{equation}
where $r_\text{yes}$ denotes the case where the response to the polar question is "yes". 

\textbf{Response Simulation Step:} Here, the response simulator is a VQA model that calculates $P(r \mid q, i)$ for any $(r,q,i)$ tuple. We then define
\begin{equation}
P(r \mid q,i)_\text{rel} \propto P(r \mid q,i) \cdot P(\text{Relevance(q,i))}
\end{equation}
where ``rel'' refers to the relevance-adjusted probability. We proceed with the information gain calculation in Eq. \ref{eq1} using $P(r \mid q,i)_\text{new}$, and the question with the highest information gain computed using the process above is then selected as the final question the model chooses to ask.

\subsection{Belief Updates}
\label{subsec:belief-updates}
The model initially has a uniform belief distribution over all images, \ie $P(i=y \mid x^0) = \frac{1}{k}$ for all $i$, where $k$ is the number of images. Recall that $x^t$ is defined as the history of the interaction $(q_1, r_1, \dots q_t, r_t)$. 
After the model asks a questions and receives a response from its partner (which is either a human or, in automatic evaluation, another VQA model --- see Section \ref{subsec:self-play}), the model will need to update its internal beliefs conditioned on the given response. As discussed in Section \ref{subsec:task}, our setting allows two types of responses, either the standard response or a null response (\ie responding with "No Answer"). This is the response-side analog to the $P(\text{Relevance}(q,i))$ discussed in \ref{subsec:top-q-selection}. It allows us to account for presuppositions, so we can properly update the model depending on whether or not an image applies to the question. We outline the two possible scenarios below:
\paragraph{Standard Response.}  
This describes the scenario when the model provides an actual answer to the question (rather than responding with ``No Answer''). Here, we wish to compute $P(y \mid x^t,q,r)$. We can apply Bayes' Rule to obtain
    \begin{equation}
    \label{eq4}
    P(y | x^t,q,r) \propto \\
    P(r|x^t,y,q)P(q|x^t,y)P(y|x^t)
    \end{equation}
where $P(r \mid x^t,y,q)$ can be calculated using the VQA model, $P(q \mid x^t,y)$ can be calculated using the question selector model, and $P(y \mid x^t)$ can be calculated recursively. Full details for these calculations can be found in \autoref{appendix:belief-updates}.

\paragraph{Null Response.} This describes the scenario when the model responds with ``No Answer''.
    Here, the calculation for $P(y \mid x^t,q,r_{null})$ is done similarly to the standard case above, except with the change that we consider $P(r_{null} \mid x^t,i_y,q)$ instead of $P(r \mid x^t,i_y,q)$. Because "No Answer" is not in the vocabulary for our VQA model, we instead design a proxy method to calculate $P(r_{null} \mid x^t,i,q)$. We thus define $P(\text{Irrelevance}(q,i))$ and calculate it similar to Eq. \ref{eq2}, but with $r_{no}$ instead of $r_{yes}$ (more details in \autoref{appendix:belief-updates}.)
    We then approximate $P(r_{null} \mid i,q) \approx P(\text{Irrelevance}(q,i))$. 
    When multiplied with the other probabilities in Eq. \ref{eq4}, this effectively results in upweighting the images which do not contain the subject, and downweighting the images which contain the subject.

The entire process is then repeated until one of the beliefs in the model's distribution over images exceeds a certain threshold $\gamma$.

%% file: sections/4.experimental-setup.tex
\section{Experimental Setup}
\label{sec:experimental-setup}
We describe our experimental setup in Sections \ref{subsec:datasets-image-selection}, \ref{subsec:modules}, and \ref{subsec:models-and-baselines}. We then compare our approach to baselines in both games with human partners (Section \ref{subsec:human-played-games}), as well as in automatic model-based evaluations (Section \ref{subsec:self-play}).

\subsection{Datasets and Image Selection}
\label{subsec:datasets-image-selection}
We compile image sets from the MS-COCO dataset \cite{mscoco}. Image sampling is done under various configurations. In the \textbf{easy} setting, $k=10$ images are simply sampled randomly from the MS-COCO validation set, as was done in \citet{white-etal-2021-open}. We refer to this setting as easy since, with sufficiently diverse images and a strong questioning model, most of these cases should be quickly solvable by simply asking a general question like ``What is the subject of the image?'' 

In order to test the question-asking ability of the model,
we also reduce the diversity among the images, \ie make the images more similar to each other. This also more closely reflects real-world retrieval settings, where a group of images may share many similarities.
To identify similar images, we parse the MS-COCO ground truth captions to identify the main subject of each image. For the \textbf{hard} setting, we select $k=10$ images which all share a single subject, while for the \textbf{medium} setting, we select 2 subjects, with $5$ images all sharing the first subject, and the remaining $5$ images sharing the second subject. 
An example of a hard setting can be found in \autoref{fig:game}.

\subsection{Modules}
\label{subsec:modules}
We previously highlighted multiple modules in our pipeline. We detail each of them below.

\paragraph{Image Captions.} To test the question-asking capabilities of our approach, we wish to make the captions as accurate as possible in order to reduce the possible sources of error in the pipeline. As such, we use the provided ground truth captions in the MS-COCO dataset.

\paragraph{Candidate Question Generation Model.} We want a strong model that can use in-context learning to convert captions to candidate questions. For this, we use GPT-4 \cite{gpt4}.

\paragraph{Question Selection Model.} For both the Existence Simulator Model and the Response Simulator Model, we use ViLT-VQA \cite{vilt}, which is a strong vision-language model trained on the VQA-v2 benchmark \cite{vqav2}.

\paragraph{User Response Simulator.} 
We consider a variety of possible sources for the ground truth. In the basic setting, we evaluate models in games with human partners (more details in Section \ref{subsec:human-played-games}). However, we also consider a self-play version, where we use a separate VQA model to serve as the responder (more details in Section \ref{subsec:self-play}). We call this responder model the user response simulator, and we use it as a proxy for the human when conducting the self-play evaluations. Specifically, we use BLIP \cite{blip} as our user response simulator.

\begin{table*}[t]
    \centering
    \begin{tabular}{lcccccc}
        \toprule
        & \multicolumn{2}{c}{Easy} & \multicolumn{2}{c}{Medium} & \multicolumn{2}{c}{Hard} \\
        \textbf{Model} & \textbf{Accuracy} & \textbf{Turns} & \textbf{Accuracy} & \textbf{Turns} & \textbf{Accuracy} & \textbf{Turns} \\
        \cmidrule(lr){1-1} \cmidrule(lr){2-3} \cmidrule(lr){4-5} \cmidrule(lr){6-7}
        Polar Yes/No & 0.78 & 3.81 & 0.72 & 3.62 & 0.69 & 3.60 \\
        Open-Ended (No Presupp.) & 0.95 & 1.38 & 0.84 & 1.99 & 0.81 & 2.74  \\
        Open-Ended (With Presupp.) & 0.93 & 1.55 & 0.85 & 1.86 & 0.83 & 2.46 \\
        \bottomrule
    \end{tabular}
    \caption{Averaged results for self-play evaluations. Each (model, difficulty) pair is evaluated over 80 games.}
    \label{tab:self-play-results}
\end{table*}

\begin{table*}[t]
    \centering
    \begin{tabular}{lcccc}
        \toprule
        & \multicolumn{2}{c}{Easy} & \multicolumn{2}{c}{Hard} \\
        \textbf{Model} & \textbf{Accuracy} & \textbf{Turns} & \textbf{Accuracy} & \textbf{Turns} \\
        \cmidrule(lr){1-1} \cmidrule(lr){2-3} \cmidrule(lr){4-5}
        Polar Yes/No & 0.73 & 3.32 & 0.68 & 3.38 \\
        Open-Ended (With Presupp.) & 0.83 & 1.70 & 0.73 & 2.73 \\
        \bottomrule
    \end{tabular}
    \caption{Averaged results for human evaluations. Each (model, difficulty) pair is evaluated over 40 games. 
    \vspace{-1em}
    }
    \label{tab:human-results}
\end{table*}

\subsection{Models and Baselines}
\label{subsec:models-and-baselines}
We test the performance of the following pipelines:
\begin{enumerate}
    \item \textbf{Polar Yes/No Questions} -- This closely follows the method of \citet{white-etal-2021-open}, but instead of their previously used CNN-based VQA classifier, we use the VQA models listed in Section \ref{subsec:modules}, as we found these off-the-shelf pretrained VQA models to perform better than training our own CNN-based classifier. 
    More details on Yes/No question generation can be found in \autoref{appendix:yesno}.
    \item \textbf{Open-Ended Questions (No Presupp.)} -- This is the pipeline described in Section \ref{sec:methods}, except that it does not contain the existence simulation step during question selection, and it does not allow "No Answer" as a response.
    \item \textbf{Open-Ended Questions (With Presupp.)} -- This is exactly the pipeline described in Section \ref{sec:methods}.
\end{enumerate}
In terms of evaluation metrics, we consider game length (number of turns) and accuracy (how often the model guesses correctly).

\subsection{Human Evaluation}
\label{subsec:human-played-games}
We recruited annotators from Amazon Mechanical Turk to play the game with the model. We selected annotators who have completed our qualification test and have >98\% acceptance rate and >10k completed HITs. For each HIT, the annotators would be paired with a random model type (polar questions versus open-ended questions) and presented with $k=10$ images, where the target image would be highlighted. The human would provide responses to the model's questions, and the game automatically stops when the model's confidence exceeds a certain threshold. For the open-ended setting, there is a separate button for "No Answer" that the user can select. More details on the human evaluation process can be found in \autoref{appendix:human}.

\subsection{Model-Based Evaluation}
\label{subsec:self-play}
We use a BLIP VQA model \cite{blip} to provide a response $r$ given $(q,I)$. This is straightforward for the polar yes/no and for open-ended (no presupp.) settings. However, for the open-ended (with presupp.) setting, we need a way to provide the ``No Answer'' response. To do this, we first convert the question $q$ into polar yes/no questions $\{q_{yn_1}, q_{yn_2}, \dots q_{yn_j}\}$. We then feed each of these $q_{yn}$ yes/no questions to BLIP to receive a yes/no response $r_{yn_j} = \arg\max_{r \in \{\text{yes,no}\}} VQA(r \mid q_{yn_j},i)$. If at least half of the responses to $\{q_{yn_1}, q_{yn_2}, \dots q_{yn_j}\}$ are ``no'', then the self-play response model responds with ``No Answer.'' Otherwise, it treats it as a standard question and provides a standard answer.

%% file: sections/5.results-analysis.tex
\section{Results}
\label{sec:results-and-analysis}

\subsection{Self-Play Evaluations}
\label{subsec:selfplay-evals}
We see in \autoref{tab:self-play-results} that the model which asks open-ended questions while explicitly handling presuppositions performs the best, both in terms of accuracy and number of turns. In general, both open-ended methods perform much better than the polar yes/no questions in terms of game length, which demonstrates that open-ended questions indeed fetch more information for the model. 

One interesting finding in the self-play experiments is that even without proper presupposition handling, the naive open-ended model already performs relatively well most of the time. We notice that the gap between the naive open-ended model and the smart open-ended model only begins to reveal itself in the medium and hard settings. This is likely because in the easy setting, the images are often diverse enough to be solvable in 1 or 2 turns, as evidenced by the low number of turns. As such, there likely will not be many presuppositions encountered early on, as a general probing question like ``What is the main subject of the image?'' would be very highly informative and be (correctly) selected as the first question a majority of the time. In contrast, when the images are more similar to each other, the games will naturally take longer, and there will be more opportunities for scenarios with presupposition errors to appear. We further verify this hypothesis in an even more challenging setting in Section \ref{subsec:ablation-hard-setting}.

\subsection{Human Evaluations}
We validate these self-play results using human evaluations, comparing the polar yes/no setting against the open-ended (with presupp.) setting. We select the easy and hard settings, and conduct 40 human-played games per (model, setting) pair. 
Results are given in \autoref{tab:human-results}.

These human evaluations corroborate our self-play findings, demonstrating that open-ended questions, when asked properly, indeed outperform the polar yes/no method of \citet{white-etal-2021-open}. There is a slight drop in human performance as compared to the self-play performance. This is likely because the VQA models used in self-play are able to capture very tiny details in the images, which humans may look past or fail to discern (e.g., ``How many [X] are there in the image?'' may be troublesome for a human to answer if $X>10$).

\subsection{Scaling To Large Image Sets}
\label{subsec:ablation-hard-setting}

In Section \ref{subsec:selfplay-evals}, we observed that the performance improvement from presupposition handling in the open-ended model begins to widen as the difficulty of the task increases. Here, we further verify this by considering an even more challenging task: increasing the number of images $k$ from $10$ to $100$, while still maintaining the hard setting of sharing the same subject.

In \autoref{tab:ablation-k}, we see an even larger improvement from presupposition handling in this more challenging setting. Notably, there is a significant accuracy drop for the open-ended (no presupp.) setting, which is likely because as these presupposition errors appear in this challenging setting, they accumulate over multiple turns and lead to worse belief updates. This does not happen in the open-ended (with presupp.) model, where accuracy stays roughly the same as the $k=10$ case. For game length, the number of turns increases for all settings beyond $k=10$, but the increase is most substantial for the polar and No Presupp settings, while remaining minimal for the open-ended setting that handles presuppositions. This demonstrates that being able to avoid presupposition errors is indeed beneficial for the model to truly be able to ask the most informative open-ended questions.

\begin{table}[t]
    \centering
    \scalebox{0.93}{
    \begin{tabular}{lcc}
        \toprule & \multicolumn{2}{c}{k=100} \\
        \textbf{Model} & \textbf{Accuracy} & \textbf{Turns} \\
        \midrule
        Polar Yes/No & 0.73 & 13.4 \\
        Open-Ended (No Presupp.) & 0.73 & 9.6 \\
        Open-Ended (With Presupp.) & 0.83 & 3.8 \\
        \bottomrule
    \end{tabular}
    }
    \caption{Averaged self-play results for the large-scale image set challenge setting where number of images $k=100$. Each (model, difficulty) pair is evaluated over 30 games.
    \vspace{-1em}
    }
    \label{tab:ablation-k}
\end{table}

%% file: sections/6.discussion.tex
\section{Analysis and Ablation Studies}
\label{sec:ablation-studies}
We conduct an ablation study (Section \ref{subsec:ablation-presupposition-handling}) to examine the importance of various components of the presupposition handling process.

\begin{table}[t]
    \centering
    \begin{tabular}{lcc}
        \toprule
        \textbf{Presupp. Handling} & \textbf{Accuracy} & \textbf{Turns} \\
        \midrule
        None (No Presupp.) & 0.81 & 2.74 \\
        None (Double Update) & 0.72 & 1.60 \\
        \midrule
        Only in question selection & 0.81 & 2.73 \\
        Only in belief updates & 0.83 & 3.54 \\
        Both (With Presupp.) & 0.83 & 2.46 \\
        \bottomrule
    \end{tabular}
    \caption{Presupposition ablation results for the open-ended questions (self-play). Here, "Double Update" refers to simply conducting the belief updates twice in each turn.
    \vspace{-1em}
    }
    \label{tab:presupposition-ablation-results}
\end{table}

\begin{table*}[t]
    \includegraphics[width=0.9\textwidth]{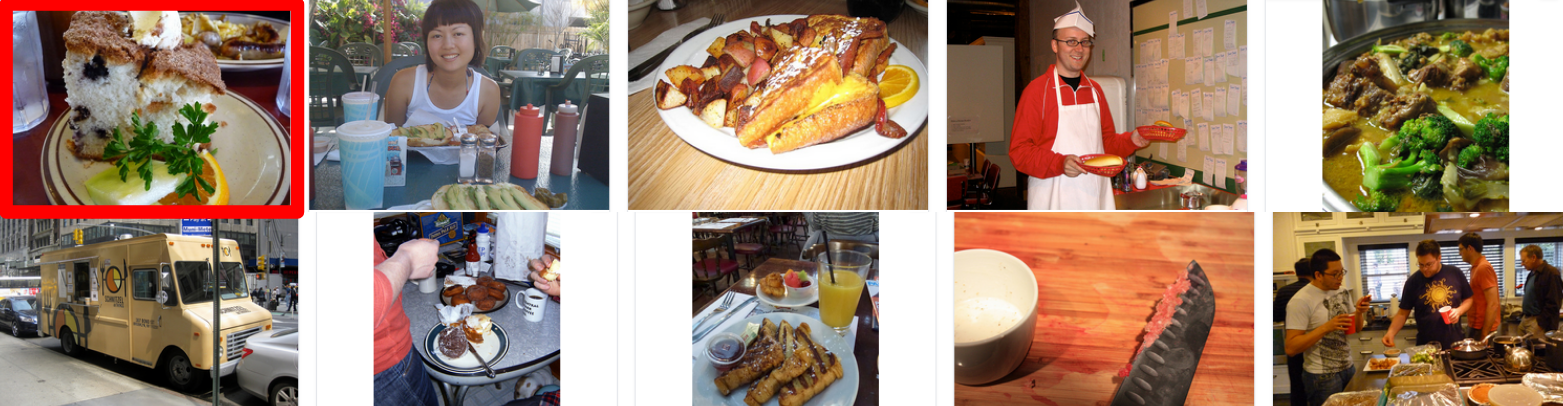}
    \centering
    \small
    \begin{tabular}{p{0.3\textwidth}p{0.4\textwidth}ll}
        \textbf{Setting} & \textbf{First Question} & \textbf{Correct?} &        \textbf{Turns} \\
        \toprule
        Polar Yes/No & Is there a restaurant table? & No & 3 \\
        Open-Ended (No Presupp.) & What is next to the knife? & No & 2 \\
        Open-Ended (With Presupp.) & What type of food are the people eating? & Yes & 1 \\
    \end{tabular}
    \caption{First question asked by each model in the game above, together with game statistics. Full game conversations can be found in \autoref{appendix:examples}.
    \vspace{-1.5em}
    }
    \label{tab:qualitative-results}
\end{table*}

\subsection{Presupposition Handling}
\label{subsec:ablation-presupposition-handling}
Because our method employs a two-step process in the belief updates, one possible explanation for the performance improvement could be an increase in the number of belief updates being performed. To test this, we consider a case where we do not do any presupposition handling, but instead simply conduct the belief updates twice in each round. This is represented by "None (Double Update)" in \autoref{tab:presupposition-ablation-results}. We observe that performing this double update results in significantly shorter games, which is likely due to each turn having sharper updates. However, the accuracy also suffers a significant drop, likely because even the incorrect beliefs will get updated twice. In contrast, our proposed method is able to reduce the number of turns without compromising the accuracy.

We further examine the effect of removing various components of the presupposition handling pipeline. Specifically, we consider what happens when we remove presupposition handling during question selection, as well as when we remove it during belief updates. As we see in \autoref{tab:presupposition-ablation-results}, only adding presupposition handling in the question selection results in very little change. This may be because the better selection quality is offset by the model's inability to update its beliefs accordingly. On the other hand, only adding it in the belief updates results in much longer turns. Qualitatively, we observed that the model tends to get "stuck" more in this scenario, as there are games which would exceed 10 turns.
These ablations confirm the importance of our full approach for handling presuppositions in informative question generation.

\subsection{Qualitative Analysis}
\autoref{tab:qualitative-results} contains an example set of images, together with the self-play results for three types of models. The images were taken from a game in the hard setting, with the common subject ``food.'' 

Here, the open-ended (with presuppp.) model performs the best, finishing the game in a single turn (Answer: ``Dessert'') and guessing the response correctly. Meanwhile, if we remove the presupposition handling, the model generates an unanswerable first question ``What is next to the knife?'', as there is no knife in the image. The self-play user simulator responds with ``cake'', but since the model cannot handle presuppositions, it mistakenly gets led towards Image 7 instead of Image 1 and hence predicts incorrectly. Meanwhile, the polar yes/no setting asks a sensible question, but it is quite inefficient compared to the type of questions asked by the open-ended questions.

\subsection{Further Discussion}
Here, we address the topic of human cognitive load. This is crucial for a system which interacts with humans, as we do not wish to increase accuracy at the expense of substantially increasing cognitive effort for users. Indeed, a part of our model's performance improvements likely emerge due to the dialogue partner offering more comprehensive information. However, we argue that our model does not lead to an increased cognitive load but rather makes it easier overall for the users. We can view total cognitive load as a variable dependent on two factors: the number of questions in the interaction, and the cognitive load-per-question. Our approach produces a substantial reduction in the number of questions required, from 3.3 turns to 1.7 turns in the easy setting and 3.4 to 2.7 turns in the hard setting (\autoref{tab:human-results}). Although investigating load-per-question is inherently difficult, we argue qualitatively that the "wh-" questions generated by our approach are typically natural (\autoref{fig:appendix-games-screenshots-2}) without very complex syntactic structure and typically involving binary relationships (e.g. "Q: What is next to the knife? A: Cake.") Indeed, when playing this game, humans naturally ask open-ended questions without placing undue burden on their communicative partners. 

%% file: sections/7.related-work.tex
\section{Related Work}
\label{sec:related-work}

\paragraph{Interactive information seeking}
Ambiguity is a persistent challenge in language.
Recent work has approached ambiguity resolution through the lens of interactive information seeking, borrowing from the optimal experimental design literature \citep{lindley}.
These methods rely on a partner model, 
which is used to measure the informativity of questions 
\citep{rao-daume-iii-2018-learning,Lee2018AnswererIQ,ingress,yu-etal-2020-interactive}.
Most related to ours is the work by \citet{white-etal-2021-open},
which proposes an unsupervised approach to informative question generation.
Their approach is limited to polar yes/no questions, which we extend to open-ended \emph{wh-} questions.
\citet{krishna2019information} also present an approach to generating open-ended informative questions by training a variational autoencoder on VQA datasets. Similar to our approach, their method selects questions that are both (1) informative and (2) relevant to the image. Our approach optimizes for similar objectives, but uses off-the-shelf VQA and LLMs without any training data for generating questions.

\paragraph{Presupposition errors and question decomposition}
Presupposition errors lead to one form of unanswerable question \citep{Davis_2020}.
Such questions have been extensively explored in literature \citep{zhu-etal-2019-learning, kim-etal-2021-linguist}.
Past work has shown that models trained on datasets without unanswerable questions often fail to generalize to unanswerable questions \citep{rajpurkar-etal-2018-know, kim-etal-2023-qa}.
We propose a method for adapting a model trained only on answerable questions to unanswerable questions via question decomposition, without supervision.
This contrasts with recent work on question decomposition, which has focused on decomposing complex questions \citep{perez-etal-2020-unsupervised,selfask}.

\paragraph{Collaborative reference games}
Collaborative reference games focus on building common ground in a symmetric dialogue setting, where both participants have equal roles
\citep{mf,photobook,onecommon,udagawa-aizawa-2021-maintaining,fried,pip}.
While both settings require reasoning, we focus on the asymmetric question-answering setting,
where the asymmetry prevents the questioner from relying too heavily on their partner.

%% file: sections/8.conclusion.tex
\section{Conclusion}
\label{sec:conclusion}

We present an approach for generating open-ended informative questions in a grounded multi-turn image identification task. As compared to previous methods which ask questions with constrained answer spaces, our method is able to ask more general questions. 
We show that directly asking open-ended questions may lead to presupposition errors, where off-the-shelf VQA models may answer questions despite their irrelevance to images. To address this, we propose a two-stage method where we first formulate a question to verify the presupposition, then update the belief distribution over the images accordingly. Through both human and self-play evaluations, we show that asking presupposition-aware open-ended questions outperforms the previous state-of-the-art in both accuracy and efficiency.

%% file: sections/acknowledgements.tex
\section*{Acknowledgements}
We thank Jing Yu Koh and Saujas Vaduguru for feedback on initial drafts of our paper. We thank Jiefu Ou for guidance with Amazon MTurk evaluations.

%% file: sections/limitations.tex
\section*{Limitations}
The method that we are proposing is able to ask questions that achieve high information gain. However, there are some limitations to the way the model generates and selects questions, as well as with the way self-play evaluations are conducted. First, the reponses allowed by the system is limited by the answer space of the VQA models. While this is usually very exhaustive (around 4000 for ViLT), it may not be enough to cover certain good answers that humans may want to give to certain questions. In addition, questions are generated using captions, not directly from images. Because of this intermediate step, there may be additional room for the models to make an error. In our medium and hard game settings, we also select similar images using a subject/caption system, rather than taking into account the images. This means that the model may miss out on certain parts of the image that the caption is not able to capture. One alternative is to instead consider using the image space using more modern models such as CLIP. Lastly, with our self-play experiments, our response models only answer based on the question and the target image. It does not take into consideration all the other images, which may be crucial in helping to provide answers that can better clarify or disambiguate.

%% file: sections/ethics.tex
\section*{Ethics}
\label{sec:ethics}
In our human evaluations, we recruit workers from Amazon Mechanical Turk. We made sure to compensate these annotators fairly, paying them $0.25$ USD per HIT. From our estimates, a HIT on average can be comfortably completed in less than a minute, so given that pace, the compensation adds up to more than $15.00$ USD per hour.

%% file: sections/appendix.tex
\section{Question Generation Prompts}
\label{appendix:prompts}
In generating the prompt to convert captions to images, we initially only used few-shot examples with a simple system prompt. However, we soon discovered that the model was generating questions which were impossible to answer just visually (e.g. ``How tall is the man?'') We thus modified our prompt. For our experiments, we settled with providing OpenAI's GPT-4 \cite{gpt4} ChatCompletion API with the following system prompt.
\begin{quote}
"You are tasked to produce reasonable questions from a given caption. \\
The questions you ask must be answerable only using visual information. \\
As such, never ask questions that involve exact measurement such as 'How tall', 'How big', or 'How far', since these cannot be easily inferred from just looking at an object. \\
Likewise, never ask questions that involve age ('How old'), composition ('What is it made of'), material, emotion, or personal relationship.\\
When asking 'Where' questions, the subject of your question must be a person or a small object.\\
Never ask questions that can be answered with yes or no. \\
When referring to objects, try to be general. For example, instead of saying 'cat', you should say 'animal'. Instead of saying 'cake', you should say 'food'. \\
I repeat, when referring to object, try to be general! \\
Good questions to ask include general 'What color', as well as general probing questions such as 'What is the man doing?" or 'What is the main subject of the image?' \\
For each caption, please generate 3-5 reasonable questions."
\end{quote}

In addition, we use the following hand-crafted few-shot examples:

\begin{enumerate}
    \item Caption: A living room with a couch, coffee table and two large windows with white curtains. Questions: What color is the couch? How many windows are there? How many tables are there? What color is the table? What color are the curtains? What is next to the table? What is next to the couch?
    \item Caption: A cat is wearing a pink wool hat. Questions: What color is the animal? What color is the hat? What is the cat wearing?
    \item Caption: A stop sign with a skeleton painted on it, next to a car. Questions: What color is the sign? What color is the car? What is next to the sign? What is next to the car? What is on the sign? Where is the car?
    \item Caption: A man brushing his teeth with a toothbrush. Questions: What is the man doing? Where is the man? What color is the toothbrush?     
\end{enumerate}

In the ChatCompletion API, these prompts are added alternately, with the user providing the caption and the assistant returning the list of questions.

\section{Belief Update Calculation}
\label{appendix:belief-updates}
In Section \ref{subsec:belief-updates}, we defined how the standard response (\ie when the human responds with an actual answer rather than ``No Answer'') and the null response (\ie when the human responds with ``No Answer'') affect the belief updates. We list down the more fine-grained details below.

\paragraph{Standard Response.}
We wish to compute the probability $P(y \mid x^t,q,r)$. Note that here, we use $y$ to denote the indicator variable that $i=y$. For all images $i$ we can apply Bayes' Rule to obtain that $P(y \mid x^t,q,r)$ is proportional to
\begin{equation}
\label{eq4-appendix}
P(r \mid x^t,y,q)P(q \mid x^t,y)P(y \mid x^t)
\end{equation}
Here, $P(r \mid x^t,y,q)$ can be calculated from the VQA model described in Section \ref{subsec:top-q-selection} (we assume $VQA(r \mid q,i)$ is independent of the history $x^t$), while $P(q \mid x^t, y)$ is essentially the result of the question selection process. Lastly, $P(y \mid x^t)$ is recursively computed since we know $P(y \mid x^0)$ (uniform) and $P(y \mid x^t) = P(y \mid x^{t-1},q,r)$ by definition. These probabilities are then multiplied and normalized to compute $P(y \mid x^t,q,r)$ for each image at each timestep $t$.

\paragraph{Null Response.}
Here, the calculation for $P(y \mid x^t,q,r_{null})$ is done similar to the standard case, except with the change that we consider $P(r_{null} \mid x^t,y,q)$ instead of a standard $r$. Because ``No Answer'' is not in the vocabulary for our VQA model, we instead design a proxy method to calculate $P(r_{null} \mid x^t,i,q)$. To do this, we once again assume that $x^t$ is independent of $VQA(r \mid q,i)$ and calculate $P(r_{null} | i,q)$ similar to how we calculated $P(\text{Relevance}(q,i))$ in Eq. \ref{eq2}. However, rather than calculating $\frac{1}{j} \sum_{k=1}^j VQA(r_\text{yes} | q_{yn_k}, i)$ as in Eq. \ref{eq2}, we instead calculate $P(\text{Irrelevance}(q,i)) = \frac{1}{j} \sum_{k=1}^j VQA(r_\text{no} | q_{yn_k}, i)$. We then approximate $P(r_{null} \mid i,q) \approx P(\text{Irrelevance}(q,i))$. When multiplied with the other probabilities in Eq. \ref{eq4}, this results in upweighting the images which do not contain the subject, and downweighting the images which contain the subject.

\section{Yes/No Question Generation}
\label{appendix:yesno}
To generate a set of yes/no questions from a given caption, we follow the method of \citet{white-etal-2021-open}. First, we use a constituency parser \cite{kitaev-klein-2018-constituency} to parse the nouns and noun phrase subtrees in the caption. These nouns and noun phrases are then directly converted to questions by asking ``Is there a [NP]'' or ``Are there [NP]'' based on rule-based plurality checking. These questions are then used as the yes/no questions generated from the caption.

\section{Additional Experiment Settings}
\label{appendix:additional-experiment-settings}
The game is continued until the model's confidence in a certain image exceeds a certain threshold $\gamma$. Based on our initial validation experiments, we found $\gamma=0.8$ to work the best. In instances where the model never reaches $\gamma$ within a certain number of turns, we simply terminate the game and select the image with the highest probability according to the model. In our experiments, we stopped the game if it exceeded 20 turns. 

\begin{table*}[t]
    \includegraphics[width=0.95\textwidth]{figures/qualitative.png}
    \centering
    \begin{tabular}{p{0.3\textwidth}p{0.4\textwidth}p{0.2\textwidth}}
        \textbf{Setting} & \textbf{First Question} & \textbf{Response} \\
        \toprule
        Polar Yes/No & Is there a restaurant table? & Yes \\
        & Is there a woman? & Yes \\
        & Is there a wooden table? & Yes \\
        & Guess: Image 8 (Incorrect) & \\
        \midrule
        Open-Ended (No Presupp.) & What is next to the knife? & Cake \\
        & What is on top of the food? & Sugar \\
        & Guess: Image 7 (Incorrect) & \\
        \midrule
        Open-Ended (With Presupp.) & What type of food are the people eating? & Dessert \\
        & Guess: Image 1 (Correct) & \\
    \end{tabular}
    \caption{Full game conversations for all settings.
    \vspace{-1.5em}
    }
    \label{fig:appendix-games-screenshots-2}
\end{table*}

\section{Human Evaluations}
\label{appendix:human}
We recruit annotators from Amazon Mechanical Turk. To filter for high-quality annotators, we restrict only to annotators with >10,000 completed tasks, >98\% acceptance rate, and are located in the United States. To further ensure that the annotators understand the requirements and rules of our tasks, we first add a qualifying requirement in the form of a test that the annotators must complete before they can perform any HIT. 

This test consists of four sample games of our task, each chosen specifically to test for a particular skill in completing our task. Specifically:
\begin{enumerate}
    \item Q1: Sample task for the polar yes/no setting. Main goal is to get the annotators familiar with this setting. Must complete in $\leq$ 3 turns.
    \item Q2: Sample task for the open-ended \emph{wh-} setting. Main goal is to get the annotators familiar with this setting. Must complete in 1 turn. 
    \item Q3: Tests for knowing when to use the ``No Answer'' option. Must complete in $\leq$ 2 turns.
    \item Q4: Tests for being able to reason pragmatically, \ie look at all the images and identify what distinguishes the target image. Must complete in 1 turn.
\end{enumerate}
To pass this qualifying test, the annotators must correctly make the model guess the target image in all 4 rounds, and they must complete each round within the allotted number of turns. For the open-ended \emph{wh-} games, in order to reduce unpredictability in uman behavior, we restricted the anwer space to the vocabulary of the VQA model (i.e. ViLT VQA \cite{vilt}), though this is not strictly necessary.

\begin{figure*}
    \centering
    \includegraphics[width=0.95\textwidth]{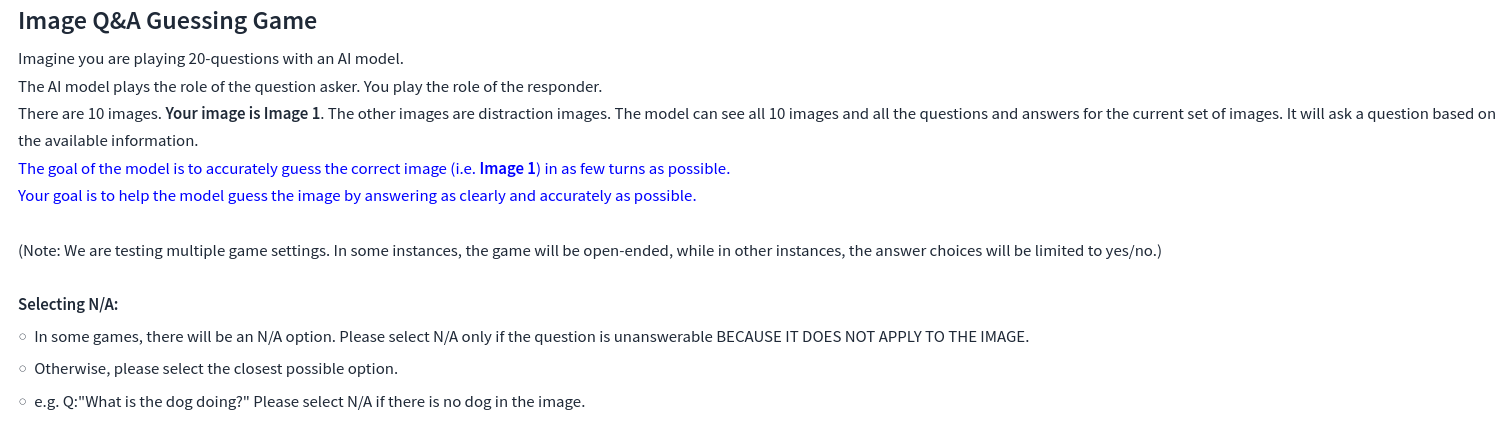}
    \includegraphics[width=0.95\textwidth]{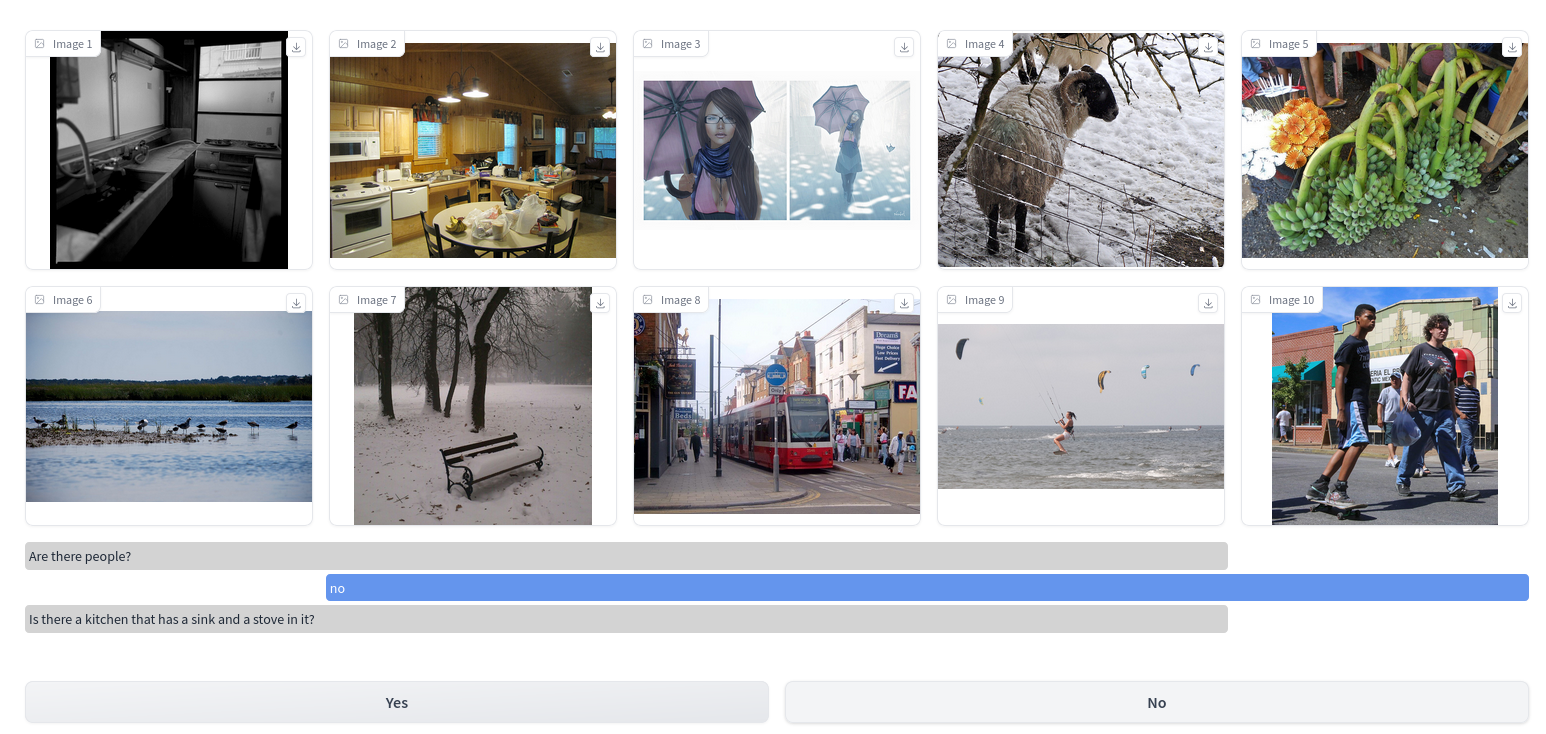}
    \includegraphics[width=0.95\textwidth]{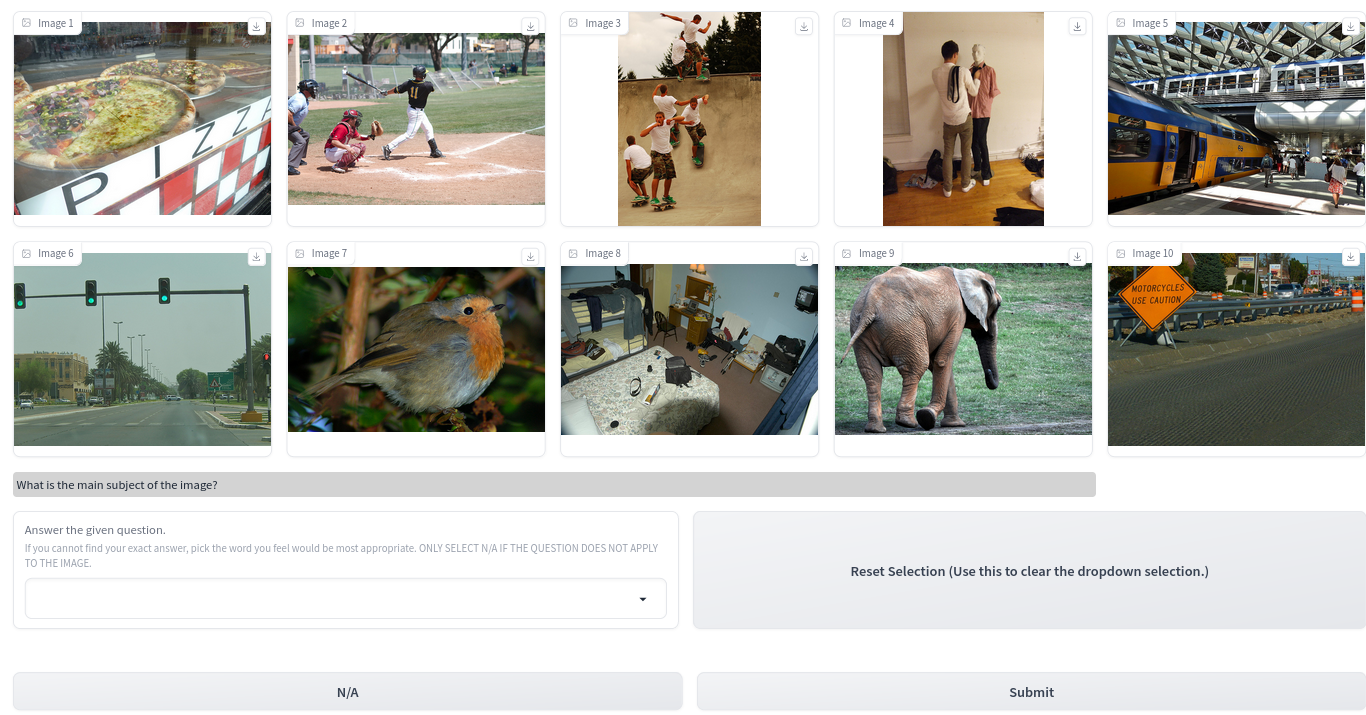}
    \caption{Our system instructions to all human annotators (top), an example of a polar yes/no game (specifically Q1 in our qualifying test) (middle), an example of an open-ended \emph{wh-} game (specifically Q2 in our qualifying test) (bottom)}
    \label{fig:appendix-games-screenshots}
\end{figure*}

We display screenshots of system instructions, as well as an example of a polar IsA game and an open-ended \emph{wh-} game in Figure \ref{fig:appendix-games-screenshots}.

We make sure to compensate our Amazon Mechanical Turk workers fairly. (See Ethics section for a discussion on this.)

\section{Additional Examples}
\label{appendix:examples}
In \autoref{tab:qualitative-results}, we outlined the game statistics for 3 different settings. In this appendix section, we list out the full games for all 3 settings in \autoref{fig:appendix-games-screenshots-2}.